\documentclass[10pt,twocolumn,letterpaper]{article}

\usepackage{wacv}
\usepackage{times}
\usepackage{epsfig}
\usepackage{graphicx}
\usepackage{amsmath}
\usepackage{amssymb}
\usepackage{booktabs}
\usepackage[table, dvipsnames, svgnames, x11names]{xcolor}
\usepackage{marvosym}
\expandafter\let\csname Cross\endcsname\relax
\usepackage{bbding}
\usepackage{array}
\usepackage{textcomp}
\usepackage{algorithm}
\usepackage{algorithmic}
\usepackage{makecell}
\usepackage{stfloats}

\def\wacvPaperID{0189} 

\wacvfinalcopy 

\ifwacvfinal
\usepackage[colorlinks,
   linkcolor=red,
   citecolor=green
]{hyperref}
\else
\usepackage[pagebackref=true,breaklinks=true,colorlinks,bookmarks=false]{hyperref}
\fi

\begin{document}

\title{Self-Supervised Monocular Depth Estimation:\\Solving the Edge-Fattening Problem}

\author{Xingyu Chen\textsuperscript{1}\qquad Ruonan Zhang\textsuperscript{1}\qquad Ji Jiang\textsuperscript{1}\qquad Yan Wang\textsuperscript{1}\qquad Ge Li\textsuperscript{1}\qquad Thomas H. Li \Letter \textsuperscript{1,2,3}\\
\small
\textsuperscript{1}School of Electronic and Computer Engineering, Peking University
\small
\textsuperscript{2}Advanced Institute of Information Technology, Peking University\\
\small
\textsuperscript{3}Information Technology R\&D Innovation Center of Peking University\\
{\tt\small \{cxy,zhangrn,jiangji,wyan\}@stu.pku.edu.cn\qquad geli@ece.pku.edu.cn\qquad tli@aiit.org.cn}\\
\url{https://github.com/xingyuuchen/tri-depth}
}

\maketitle
\thispagestyle{empty}

\begin{abstract}
   Self-supervised monocular depth estimation (MDE) models universally suffer from the notorious edge-fattening issue.
   Triplet loss, as a widespread metric learning strategy, has largely succeeded in many computer vision applications.
   In this paper, we redesign the patch-based triplet loss in MDE to alleviate the ubiquitous edge-fattening issue.
   We show two drawbacks of the raw triplet loss in MDE and demonstrate our problem-driven redesigns.
   First, we present a \textit{min.} operator based strategy applied to all negative samples, to prevent well-performing negatives sheltering the error of edge-fattening negatives.
   Second, we split the anchor-positive distance and anchor-negative distance from within the original triplet, which directly optimizes the positives without any mutual effect with the negatives.
   Extensive experiments show the combination of these two small redesigns can achieve unprecedented results:
   Our powerful and versatile triplet loss not only makes our model outperform all previous SoTA by a large margin, but also provides substantial performance boosts to a large number of existing models, while introducing no extra inference computation at all.
\end{abstract}

\mathchardef\mhyphen="2D
\definecolor{aug_our_method_color}{RGB}{241,199,245}
\definecolor{X_color}{RGB}{255,0,0}
\definecolor{checkmark_color}{RGB}{0,176,80}
\definecolor{shallow_grey}{RGB}{220,220,220}

\section{Introduction}\label{sec:intro}

Estimating how far each pixel in the image is away from the camera is a fundamental problem in 3D computer vision.
This technique is desirable in various fields, such as autonomous driving~\cite{wang2019pseudo,you2019pseudo}, AR~\cite{luo2020consistent} and robotics~\cite{griffin2020video}.
The majority of these applications adopt the appealing off-the-shelf hardware, \textit{e.g.} Lidar sensor or RGB-D cameras, to enable agents.
On the contrary, monocular videos or stereo pairs are much easier to obtain.
Hence, quantities of researches~\cite{zhou2017unsupervised,godard2019digging,watson2019self} have been conducted to estimate promising dense depth map from only a single RGB image.
Most of them and their follow-ups formulated the problem as image reconstruction~\cite{godard2019digging,zhou2017unsupervised,watson2019self,watson2021temporal,jung2021fine}.
In particular, given a target image, the network infers its pixel-aligned depth map.
Next, with a known or estimated camera ego-motion, every pixel in the target image can be reprojected to the reference image(s) which is taken from different viewpoint(s) of the same scene.
The reconstructed image can be generated by sampling from the source image, and the training loss is based on the photometric distance between the reconstructed and target image.
In this way, the network is trained under self-supervision.

Nevertheless, these approaches suffered severally from the notorious `edge-fattening' problem, where the objects' depth predictions are always `fatter' than themselves.
We visualize the problem and analyse its cause in Fig.~\ref{fig:edge-fattening-problem}.
Disappointingly, there is no corresponding one-size-fits-all solution, let alone a lightweight one.

\begin{figure}[htb]
    \centering\includegraphics[width=0.95\columnwidth]{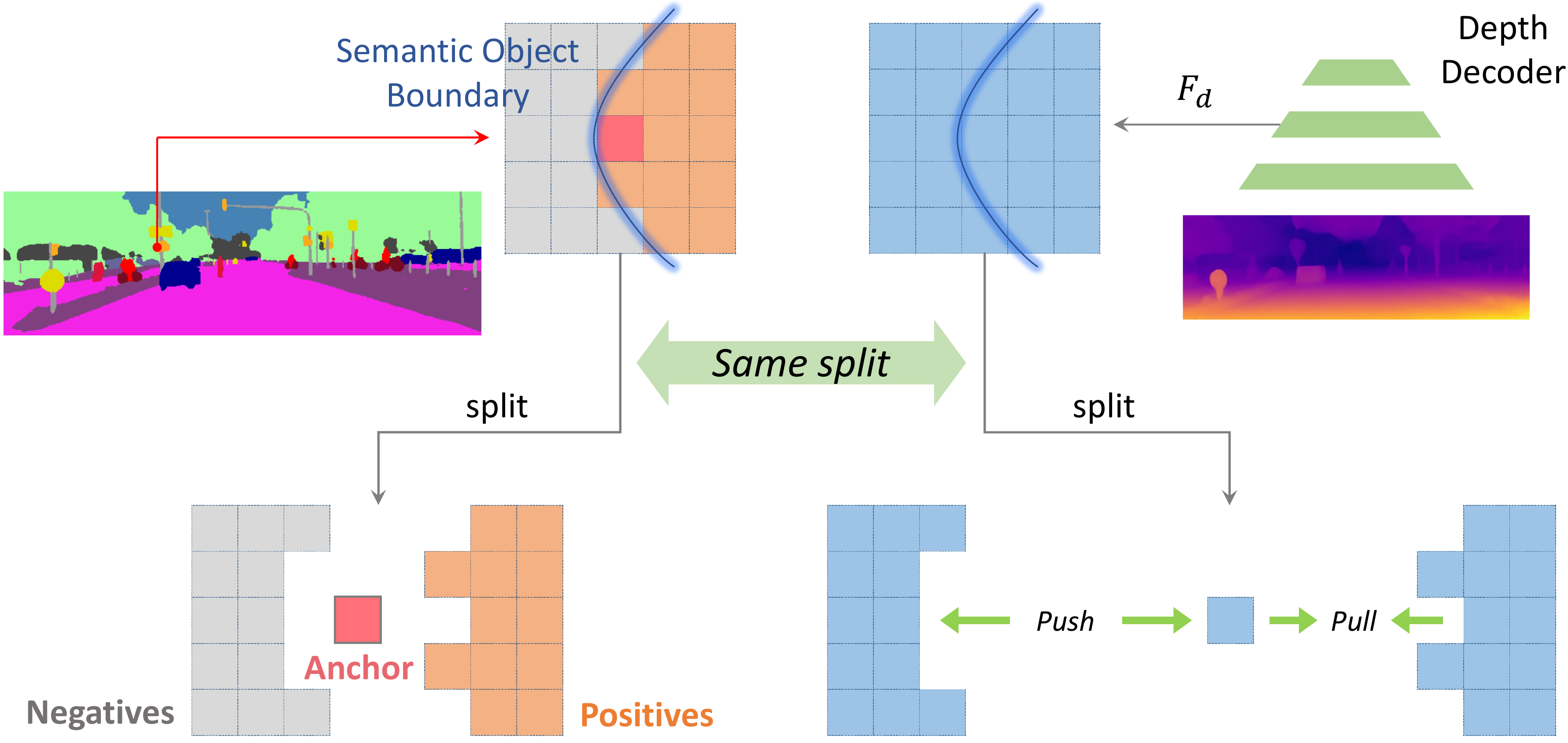}
    \caption{\textbf{The baseline semantic-aware triplet loss.} For each pixel in the semantic boundary region, we group the local patch of its corresponding depth feature ($F_d$) into a triplet according to the semantic patch. Next, the triplet loss minimizes anchor-positive distance ($\mathcal{D}^+$) and maximizes the anchor-negative distance ($\mathcal{D}^-$).}
    \label{fig:sgt}
\end{figure}

Deep metric learning seeks to learn a feature space where semantically similar samples are mapped to close locations, while semantically dissimilar samples are mapped to distant locations.
Pioneeringly,~\cite{jung2021fine} first introduced the patch-based semantics-guided triplet loss into MDE.
Its key idea is to encourage pixels within each object instance to have similar depths, while those across semantic boundaries to have depth differences, as shown in Fig.~\ref{fig:sgt}.

However, we find that the straightforward application of the triplet loss only produces poor results.
In this paper, we dig into the weakness of the patch-based triplet loss, and improve it through a problem-driven manner, reaching unprecedented performances.

First, in some boundary regions the edge-fattening areas could be thin, their contributions are small compared to the non-fattening area, in which case the thin but defective regions' error could be covered up.
Therefore, we change the optimizing strategy to only focus on the fattening area.
The problematic case illustrated in Fig.~\ref{fig:compare-avg-min-triplet} motivates our strategy, where we leave the normal area alone, and concentrate the optimization on the poor-performing negatives.

Second, we point out that the training objective of the original triplet loss~\cite{schroff2015facenet} is to distinguish / discriminate, the network only has to make sure the correct answer's score (the anchor-positive distance $\mathcal{D}^+$) wins the other choices' scores (the anchor-negative distances $\mathcal{D}^-$) by a predefined margin, \textit{i.e}, $\mathcal{D}^- - \mathcal{D}^+ > m$, while the absolute value of $\mathcal{D}^+$ is not that important, see an example in Fig.~\ref{fig:compare_cls_regres}.
However, depth estimation is a regression problem since every pixel has its unique depth solution.
Here, we have no idea of the exact depth differences between the intersecting objects, thus, it is also unknown how much $\mathcal{D}^-$ should exceed $\mathcal{D}^+$.
But one thing for sure is that, in depth estimation, the smaller the $\mathcal{D}^+$, the better, since depths within the same object are generally the same.
We therefore split $\mathcal{D}^-$ and $\mathcal{D}^+$ from within the original triplet and optimize them in isolation, where either error of the positives or negatives will be penalized individually and more directly.
The problematic case illustrated in Fig.~\ref{fig:need-of-contrastive} motivates this strategy, where the negatives are so good enough to \textit{cover up} the badness of the positives.
In other words, even though $\mathcal{D}^+$ is large and needs to be optimized, $\mathcal{D}^-$ already exceeds $\mathcal{D}^+$ more than $m$, which hinders the optimization for $\mathcal{D}^+$.

To sum up, this paper's contributions are:
\begin{itemize}
    \item We show two weaknesses of the raw patch-based triplet optimizing strategy in MDE, that it (\textit{\romannumeral1}) could miss some thin but poor fattening areas; (\textit{\romannumeral2}) suffered from mutual effects between positives and negatives.
    \item To overcome these two limitations, (\textit{\romannumeral1}) we present a \textit{min.} operator based strategy applied on all negative samples, to prevent the good negatives sheltering the error of poor (edge-fattening) negatives;
    (\textit{\romannumeral2}) We split the anchor-positive distance and anchor-negative distance from within the original triplet, to prevent the good negatives sheltering the error of poor positives.
    \item Our redesigned triplet loss is powerful, generalizable and lightweight:
    Experiments show that it not only makes our model outperform all previous methods by a large margin, but also provides substantial boosts to a large number of existing models, while introducing no extra inference computation at all.
\end{itemize}

\section{Related Work}\label{sec:related_work}

\subsection{Self-Supervised Monocular Depth Estimation}\label{subsec:self-supervised-monocular-depth-estimation}

Garg \textit{et al.}~\cite{garg2016unsupervised} firstly introduced the novel concept of estimating depth without depth labels.
Then, SfMLearner presented by Zhou \textit{et al.}~\cite{zhou2017unsupervised} required only monocular videos to predict depths, because they employed an additional pose network to learn the camera ego-motion.
Godard \textit{et al.}~\cite{godard2019digging} presented \textit{Monodepth2}, with surprisingly simple methods handling occlusions and dynamic objects in a non-learning manner, both of which add no network parameters.
Multiple works leveraged additional supervisions, \textit{e.g.} estimating depth with traditional stereo matching method~\cite{watson2019self,Tosi_2019_CVPR} and semantics~\cite{jiao2018look}.
HR-Depth ~\cite{lyu2020hr} proved that higher-resolution input images can reduce photometric loss with the same prediction.
Manydepth~\cite{watson2021temporal} proposed to make use of multiple frames available at test time and leverage the geometric constraint by building a cost volume, achieving superior performance.
~\cite{ramamonjisoa2021single} integrated wavelet decomposition into the depth decoder, reducing its computational complexity.
Some other recent works estimated depths in more severe environments, \textit{e.g.} indoor scenes~\cite{Ji_2021_ICCV} or in nighttime~\cite{vankadari2020unsupervised,Lu_2021_WACV}.
Innovative loss functions were also developed, \textit{e.g.} constraining 3D point clouds consistency~\cite{hirose2021plg}.
To deal with the notorious edge-fattening issue as illustrated in Fig.~\ref{fig:edge-fattening-problem}, most existing methods utilized an occlusion mask~\cite{NEURIPS2020_951124d4,zhu2020edge} in the way of removing the incorrect supervision under photometric loss.
We argue that although this exclusion strategy works to some extent, the masking technique could prevent these occluded regions from learning, because no supervision exists for them any longer.
In contrast, our triplet loss closes this gap by providing additional supervision signals directly to these occluded areas.

\subsection{Deep Metric Learning}\label{subsec:triplet-loss-function}
The idea of comparing training samples in the high-level feature space~\cite{chopra2005learning,bromley1993signature} is a powerful concept, since there could be more task-specific semantic information in the feature space than in the low-level image space.
The contrastive loss function (\textit{a.k.a.} discriminative loss)~\cite{hadsell2006dimensionality} is formulated by whether a \textit{pair} of input samples belong to the same class.
It learns an embedding space where samples within the same class are close in distance, whereas unassociated ones are farther away from each other.
The triplet loss~\cite{schroff2015facenet} is an extension of contrastive loss, with three samples as input each time, \textit{i.e.} the anchor, the positive(s) and the negative(s).
The triplet loss encourages the anchor-positive distance to be smaller than the anchor-negative loss by a margin $m$.
Generally, the triplet loss function was introduced to face recognition and re-identification~\cite{schroff2015facenet,zhuang2016fast}, image ranking~\cite{wang2014learning}, to name a few.
Jung \textit{et al.}~\cite{jung2021fine} first plugged triplet loss into self-supervised MDE, guided by another pretrained semantic segmentation network.
In our experiments, we show that without other contributions in~\cite{jung2021fine}, the raw semantic-guided triplet loss only yielded a very limited improvement.
We tackle various difficulties when plugging triplet loss into MDE, allowing our redesigned triplet loss outperforms existing ones by a large margin, with unparalleled accuracy.

\section{Analysis of the Edge-fattening Problem}\label{sec:problem-of-edge-fattening-issue}

\definecolor{color_p}{RGB}{90,126,201}
\definecolor{color_q}{RGB}{175,129,47}
\definecolor{color_r}{RGB}{219,50,81}
\begin{figure*}[htb]
    \centering\includegraphics[width=0.95\textwidth]{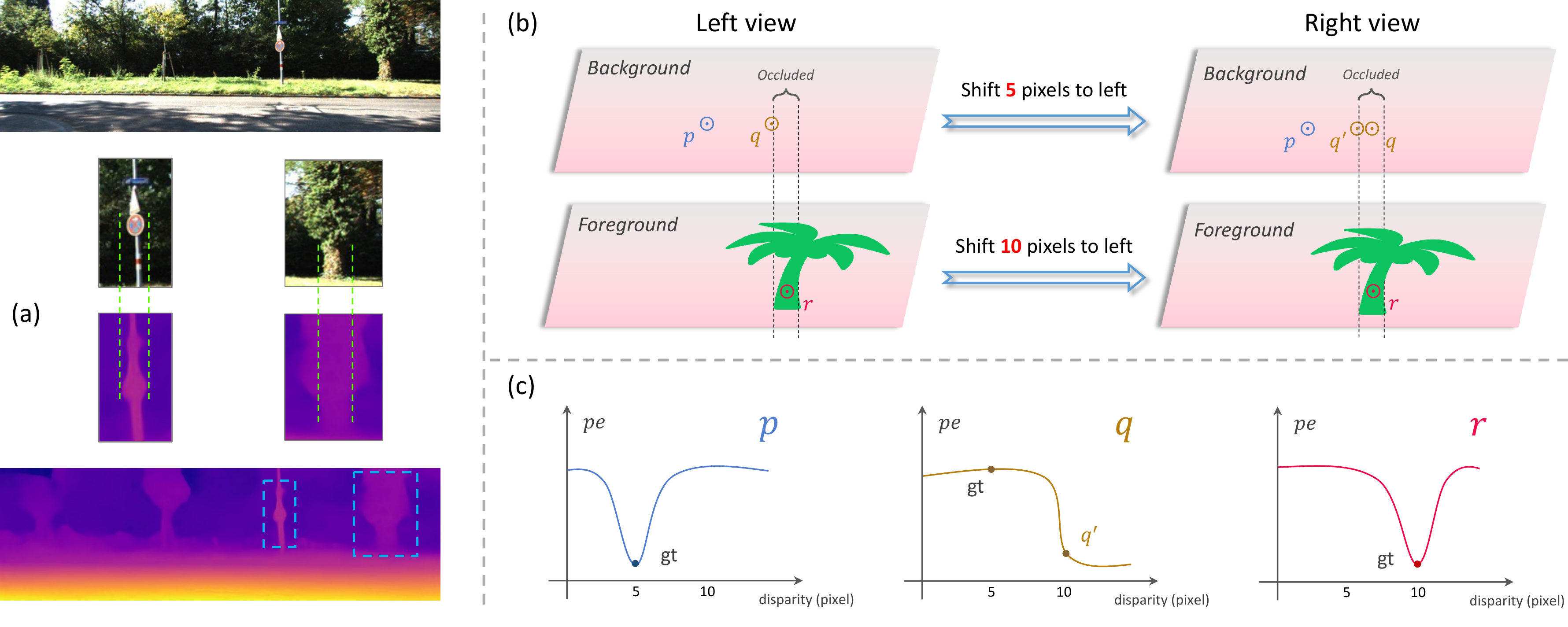}
    \caption{\textbf{Analysis of the edge-fattening issue.} \textbf{(a)} Example of the edge-fattening issue. The depth predictions of foreground objects (\textit{e.g.} the tree-trunk and poles) are `fatter' than the objects themselves. \textbf{(b)} In the left view, pixel \textcolor{color_p}{$p$} and \textcolor{color_q}{$q$} are located in the background with a disparity of 5 pixels
    , and \textcolor{color_q}{$q$} will be occluded if any further to the right. \textcolor{color_r}{$r$} is on the tree with a disparity of 10 pixels. \textbf{(c)} \textcolor{color_p}{$p$} and \textcolor{color_r}{$r$} are OK - their \textit{gt} disparity is the global optimum of the photometric error (loss). \textcolor{color_q}{$q$} suffers from edge-fattening issue. Since \textcolor{color_q}{$q$} is occluded by the tree in the right view, the photometric error of its \textit{gt} disparity 5 is large. The photometric loss function therefore struggles to find another location that has a small loss, \textit{i.e.}, shifting another 5 pixels to reach the nearest background pixel \textcolor{color_q}{$q^{\prime}$}. However, \textcolor{color_q}{$q^{\prime}$} is not the true correspondence of \textcolor{color_q}{$q$}. As a result, disparity of the background \textcolor{color_q}{$q$} equals to that of the foreground \textcolor{color_r}{$r$}, leading to the edge-fattening issue. Details in Sec.~\ref{sec:problem-of-edge-fattening-issue}.}
    \label{fig:edge-fattening-problem}
\end{figure*}

Before delving into our powerful triplet loss, it is necessary to make the problem clear.
We first show the behaviour of the so-called edge-fattening issue in self-supervised MDE, then analyse its cause in Fig.~\ref{fig:edge-fattening-problem}.
This motivates us to introduce our resigned triplet loss.

The ubiquitous edge-fattening problem, limiting performances of vast majorities of self-supervised MDE models~\cite{godard2019digging,watson2021temporal,watson2019self,jung2021fine,zhou2017unsupervised,lyu2020hr}, manifests itself as yielding inaccurate object depths that partially leak into the background at object boundaries, as shown in Fig.~\ref{fig:edge-fattening-problem}a.

We first lay out the final conclusion: \textit{The networks misjudge the background near the foreground as the foreground, so that the foreground looks fatter than itself}.

The cause could be traced back to occlusions of background pixels as illustrated in Fig.~\ref{fig:edge-fattening-problem}b\&c.
The background pixels visible in the target (left) image but invisible in the source (right) image suffer from incorrect supervision under the photometric loss, since no exact correspondences in the source image exist for them at all.

The crux of the matter is that, for the occluded pixel (\textit{e.g.} pixel $q$ in Fig.~\ref{fig:edge-fattening-problem}b), the photometric loss still needs to seek a pixel with a similar appearance (in the right view) to be its fake correspondence.
Generally, for a background pixel, only another background pixel could show a small photometric loss, so the problem turns into finding the nearest background pixel for the occluded pixel ($q$) in the source (right) view.
Since the foreground has a smaller depth $Z$ than the background, it has a larger disparity $d$ owing to the geometry projection constrain: $Z=\frac{f\cdot b}{d}$, where $b$ is the fixed camera baseline.
Consequently, the photometric loss has to shift more to the left to find the nearst background pixel (the fake solution, \textit{e.g.} pixel $q^{\prime}$ in Fig.~\ref{fig:edge-fattening-problem}b\&c).
In this way, the occluded background pixels share the same disparities as the foreground, forming the edge-fattening issue.

\section{Methodology}\label{sec:method}

\subsection{Self-supervised Depth Estimation}\label{subsec:unsupervised-depth-estimation}

Following~\cite{zhou2017unsupervised,godard2019digging}, given a monocular and/or stereo video, we first train a depth network $\psi_{depth}$ consuming a single target image $I_t$ as input, and outputs its pixel-aligned depth map $D_t = \psi_{depth}(I_t)$.
Then, we train a pose network $\psi_{pose}$ taking temporally adjacent frames as input, and outputs the relative camera pose $T_{t \rightarrow t+n} = \psi_{pose}(I_t, I_{t+n})$.

Suppose we have access to the camera intrinsics $K$, along with $D_t$ and $T_{t \rightarrow t+n}$, we warp $I_{t+n}$ into $I_t$ to generate the reconstructed image $\tilde{I}_{t+n}$:
\begin{equation}\label{eq:reconstruct}
\tilde{I}_{t+n} = I_{t+n} \big \langle proj\left( D_t, T_{t \rightarrow t+n}, K \right) \big \rangle,
\end{equation}
where $\langle \cdot \rangle$ is the differentiable bilinear sampling operator according to~\cite{godard2019digging}.

To make use of multiple input frames, we build a cost volume~\cite{watson2021temporal} using discrete depth values from a predefined range $[d_{\min}, d_{\max}]$.
Moreover, $d_{\min}$ and $d_{\max}$ are dynamically adjusted during training to find the best scale~\cite{watson2021temporal}.

In order to evaluate the reconstructed images $\tilde{I}_{t+n}$, we adopt the edge-aware smoothness loss~\cite{heise2013pm} and the photometric reprojection loss measured by $\mathcal{L}_1 + \mathcal{L}_{ssim}$~\cite{zhou2017unsupervised,godard2019digging}:
\begin{equation}\label{eq:pe}
\mathcal{L}_{pe} = \frac{\alpha}{2}(1-SSIM(I_t, \tilde{I}_{t+n})) + (1 - \alpha) \parallel I_{t} - \tilde{I}_{t+n} \parallel_1,
\end{equation}
where $\alpha = 0.85$ by default and $SSIM$~\cite{wang2004image} computes pixel similarity over a $3\times 3$ window.
See~\cite{watson2021temporal} for more details of our network architecture.

\subsection{Redesigned Triplet Loss}\label{subsec:redesigned-triplet-loss}

We first introduce our baseline depth-patch-based triplet loss, and then give detailed analyses of our two redesigns.

\subsubsection{Patch-based Triplet Loss}

\definecolor{grey_plane}{RGB}{101,101,101}
\definecolor{green_plane}{RGB}{77,159,106}
\definecolor{yellow_plane}{RGB}{148,109,83}
\begin{figure*}[t]
    \centering\includegraphics[width=0.9\textwidth]{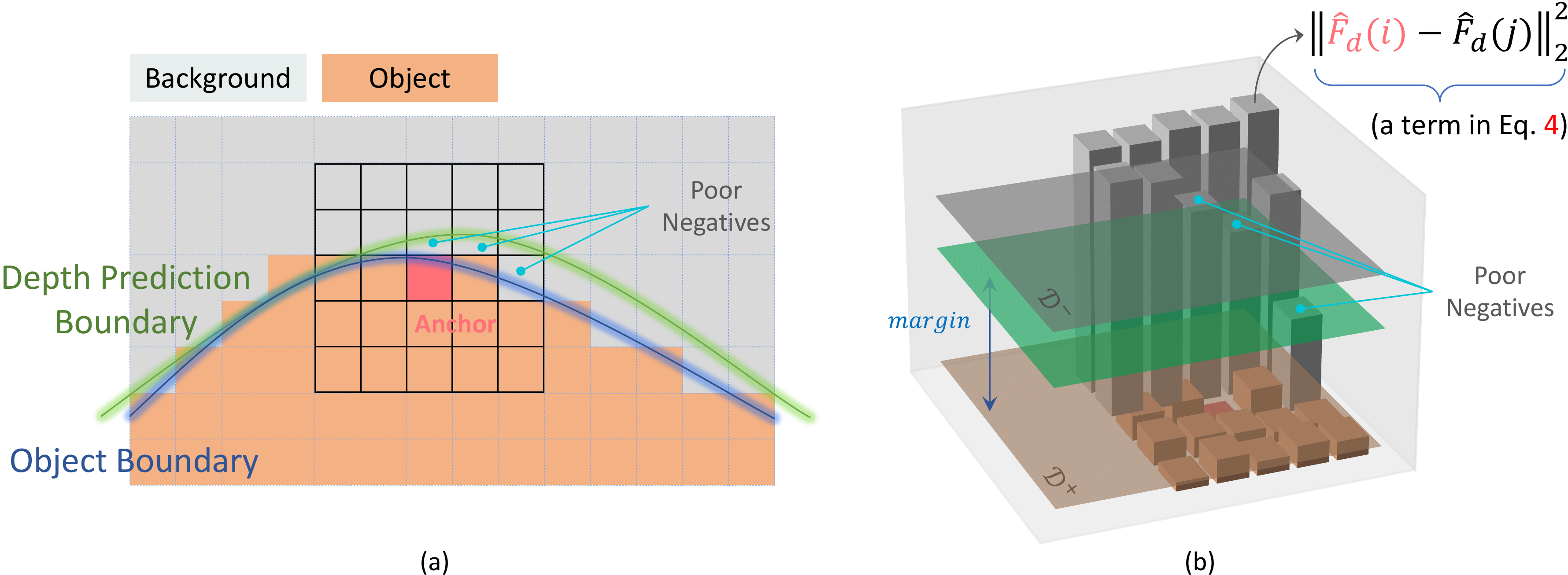}
    \caption{\textbf{How the error of the low-proportion poor negatives is sheltered by the large-proportion well-performing negatives.} \textbf{(a)} An instance of the edge-fattening issue (detailed analysis in Fig.~\ref{fig:edge-fattening-problem}), where the depth prediction is `fatter' than the object itself. In other words, negatives in the `fatter' area perform poorly - their depth has to be the same as the background but not the foreground object. \textbf{(b)} The Euclidean distance of depth feature between all pixels in the local patch and the anchor. The \textcolor{yellow_plane}{yellow plane} indicates $\mathcal{D}^{+}$ (mean anc-pos distances); the \textcolor{grey_plane}{grey plane} indicates $\mathcal{D}^{-}$ (mean anc-neg distances); the \textcolor{green_plane}{green plane} specifies the decision boundary of whether this triplet participates in training (see the hinge function in Eq.~\ref{eq:base_triplet}). $\mathcal{D}^-$ is large such that the green plane lies beneath the grey. Thus, no learning happens. Disappointingly, the poor negative pixels in the `fatter' area  (\textit{e.g.} the \textit{poor negatives} marked in (a)) get no optimization. This is because of the \textit{average} operator - the low-proportion poor negatives' contributions are weakened by the large-proportion good negatives.}
    \label{fig:compare-avg-min-triplet}
\end{figure*}

There lies a fact that pixels within a certain object have similar depths, while pixels across object boundaries may have large depth differences.
However, in many cases, the depth boundary does not align to the semantic boundary as shown in Fig.~\ref{fig:edge-fattening-problem}.
Therefore, similar to~\cite{jung2021fine}, we overcome this problem through deep metric learning~\cite{kim2020proxy,yi2014deep}.
Specifically, we group the pixels in a local patch into triplets: the central pixel as the \textit{anchor}; those share the same semantic class with the anchor as \textit{positives}; while those have a different semantic class from the anchor as \textit{negatives}.
Then, we refer to the sets of positive and negative pixels of the anchor $\mathcal{P}_i$ as $\mathcal{P}^+_i$ and $\mathcal{P}^-_i$, respectively.
For example, $|\mathcal{P}^-_i|=0$ implies $\mathcal{P}_i$ is located inside an object.

The anchor-positive distance $\mathcal{D}^+$ and anchor-negative distance $\mathcal{D}^-$ are defined as the mean of Euclidean distance of $L_2$ normalized depth features~\cite{jung2021fine}:
\begin{equation}\label{eq:depth_distance_pos}
    \mathcal{D}^+ \left( i \right) = \frac{1}{| \mathcal{P}^+_i |} \sum_{j \in \mathcal{P}^+_i} \left\| \hat{F}_d\left( i \right) - \hat{F}_d\left( j \right) \right\|^2_2,
\end{equation}
\begin{equation}\label{eq:depth_distance_neg}
    \mathcal{D}^- \left( i \right) = \frac{1}{| \mathcal{P}^-_i |} \sum_{j \in \mathcal{P}^-_i} \left\| \hat{F}_d\left( i \right) - \hat{F}_d\left( j \right) \right\|^2_2,
\end{equation}
where $\hat{F}^d = F_d / \left\| F_d \right\|$.

Intuitively, $\mathcal{D}^{+}$ should be minimized, whereas $\mathcal{D}^{-}$ should be maximized.
However, naively maximizing $\mathcal{D}^{-}$ as large as possible is unfavourable, since the depth differences between two spatially adjacent objects are not always large.
Hence, a margin $m$ is employed to prevent $\mathcal{D}^-$ from exceeding $\mathcal{D}^+$ immoderately~\cite{schroff2015facenet}.
In this way, we want:
\begin{equation}\label{eq:expect}
\mathcal{D}^{+} + m < \mathcal{D}^{-},
\end{equation}
where $m$ is the threshold controlling the least distance between $\mathcal{D}^{+}$ and $\mathcal{D}^{-}$.

In addition, since we also adopt an off-the-shelf pretrained semantic segmentation model~\cite{zhu2019improving} whose predictions could also be inaccurate, we only optimize the boundary anchors with $|\mathcal{P}^+_i|$ and $|\mathcal{P}^-_i|$ both larger than $k$~\cite{jung2021fine}:
\begin{equation}\label{eq:expect_all}
    \mathcal{D}^{+}\left( i \right) + m < \mathcal{D}^{-}\left( i \right), ~\forall~\mathcal{P}_i \in \Gamma,
\end{equation}
\begin{equation}\label{eq:boundary}
    \Gamma = \left\{ \mathcal{P}_i ~\Big|~ \left(|\mathcal{P}^+_i| > k\right) \wedge \left( |\mathcal{P}^-_i| > k \right) \right\},
\end{equation}
where $\Gamma$ is the set holding all semantic boundary pixels that fulfill the aforementioned constraint.

Consequently, the baseline triplet loss is defined as:
\begin{equation}\label{eq:base_triplet}
    \mathcal{L}_{tri} = \frac{1}{|\Gamma|} \sum_{\mathcal{P}_i \in \Gamma} \left[\mathcal{D}^{+}(i) - \mathcal{D}^{-}(i) + m \right]_{+},
\end{equation}
where $[\cdot]_{+}$ is the hinge function.

In this way, we do not require any ground truth annotations, neither in depth nor segmentation, which enables a fully self-supervised setting and can therefore better generalize to unseen scenarios.

\subsubsection{Motivation 1: The Poor Negatives' Error may be Sheltered}

Nevertheless, naively encouraging the \textit{mean} $\mathcal{D}^{-}$ in the local patch to increase leads to poor results.
We trace its cause to the average operator in computing the anchor-negative distance.
An example is demonstrated in Fig.~\ref{fig:compare-avg-min-triplet}, where the majorities of the negatives perform well, while only a few pixels inside the gap between the depth and semantic boundary perform poorly, triplet loss with the strategy of averaging $\mathcal{D}^-$ is likely to mask the error of these poor negatives, as their contributions are small compared to large numbers of other well-performed negatives.
As a result, the average anchor-negative distance is still not small enough to be optimized by the triplet loss with margin $m$, which further hinders the optimization for these thin but poor negatives.

\subsubsection{Solution 1: Focus only on the Fattening Area}

We point out that it is not a wise idea to simply increase the margin $m$ to let cases like Fig.~\ref{fig:compare-avg-min-triplet} participate in training, because it is not only these cases matter, we should find out the right $m$ suitable for as many cases as possible.
Instead, inspired by \textit{PointNetVLAD}~\cite{uy2018pointnetvlad}, a point cloud retrieval network, we introduce selecting the hardest negative to participate in training, while leaving other negatives alone.
In this way, the anchor-negative distance becomes:
\begin{equation}\label{eq:depth_distance_hardest_neg}
    \mathcal{D}^{-\prime} \left( i \right) = \min_{j \in \mathcal{P}^-_i} \left\| \hat{F}^d\left( i \right) - \hat{F}^d\left( j \right) \right\|^2_2.
\end{equation}
This comes from the fact that if the worst negative is good enough, other negatives from the local patch would no longer need optimizations.
Compared to other computer vision tasks like image classification~\cite{yu2018correcting}, point cloud retrieval~\cite{uy2018pointnetvlad}, person re-identification~\cite{hermans2017defense}, where hard-negative mining is expensive because the whole set of training data has to be revisited to compute the one with the largest $\mathcal{D}^{-}$, our strategy is much more efficient since our mining process happens only in the local patch, and it is quite easy to proceed all triplets in batches.

\subsubsection{Motivation 2: The Positives' Error may be Sheltered}

\begin{figure}[htb]
    \centering\includegraphics[width=0.95\columnwidth]{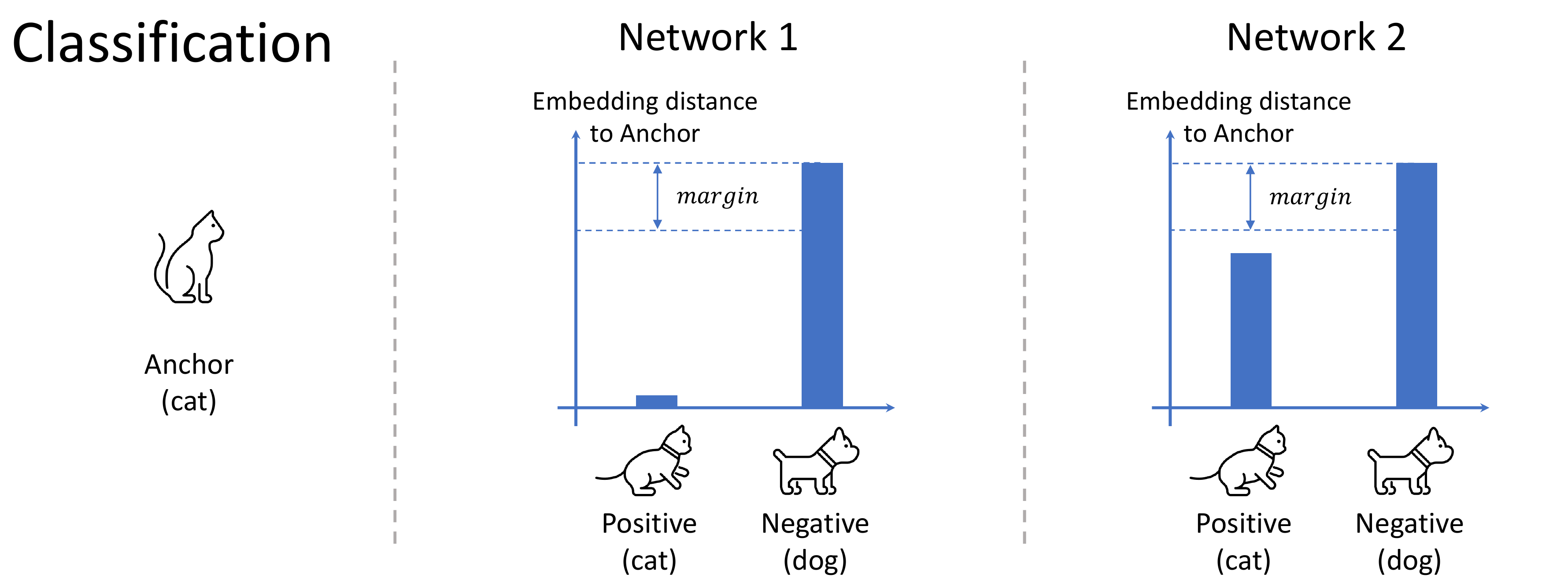}
    \caption{In a classification task, both network 1\&2 perform well (both \textit{s.t.} Eq.~\ref{eq:expect}), they undoubtedly judge the anchor as a cat because there is a clear distance gap ($> margin$) between the two choices.}
    \label{fig:compare_cls_regres}
\end{figure}

In a classification task, it is OK (tolerable) that the anchor-positive distance ($\mathcal{D}^+$) is a bit large (\textit{e.g.} Network 2 in Fig.~\ref{fig:compare_cls_regres}), because it only has to be smaller enough than any of the samples belonging to a different class (\textit{e.g.} the dog).
In other words, for classification, it is the \textit{comparison relationship} that $\mathcal{D}^- \gg \mathcal{D}^+$ counts, and neither the individual $\mathcal{D}^-$ nor $\mathcal{D}^+$ counts.
A classification network only has to make sure that the score of the correct answer wins all the others, while the absolute score of the correct answer is not that important.
In fact, it makes no difference for $\mathcal{D}^+$ to be any value $\in \left[0, \mathcal{D}^--m\right)$ in Fig.~\ref{fig:compare_cls_regres} when inferring.

\begin{figure}[htb]
    \centering\includegraphics[width=0.95\columnwidth]{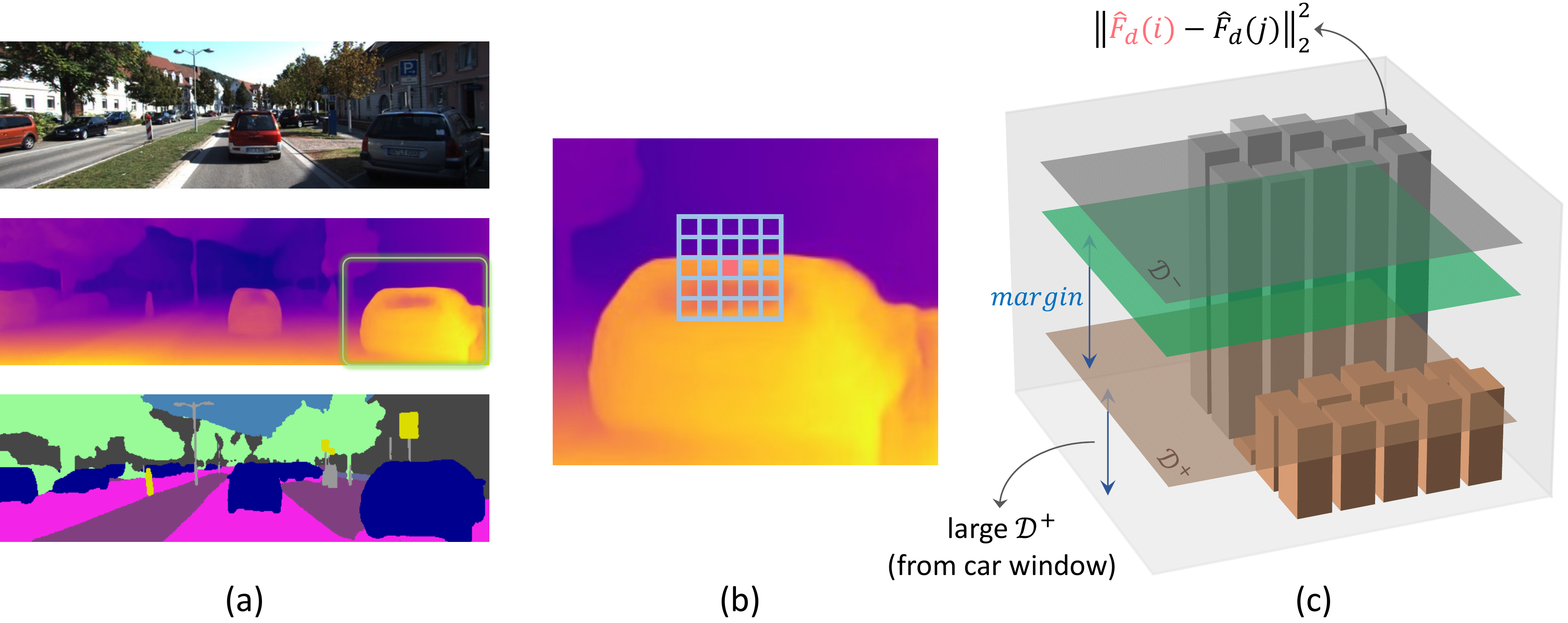}
    \caption{\textbf{Example of how the error of poor positives is sheltered by the good negatives.} \textbf{(a)} From top to bottom: An RGB image; its predicted depth; semantic segmentation. \textbf{(b)} Image patch in (a), where the depth prediction of the car window (positives) is wrong - it has to be the same as the car body, while depth predictions of the background (negatives) are extremely accurate. \textbf{(c)} The average of anchor-negative distances ($\mathcal{D}^-$) is so large that the \textcolor{grey_plane}{grey plane} lies above the \textcolor{green_plane}{green plane}. That is, the triplet loss stops working as Eq.~\ref{eq:expect} is satisfied. However, all poor positive car window pixels (the large $\mathcal{D}^+$) get no optimization.}
    \label{fig:need-of-contrastive}
\end{figure}

However, when it comes to depth estimation, the situation changes.
We emphasize that depth estimation is a regression problem, which aims to predict accurate depth values for every pixel.
In this case, $\mathcal{D}^-$ and $\mathcal{D}^+$ matter individually.
For example, as depicted in Fig.~\ref{fig:need-of-contrastive}b, the depth prediction of the car window (positives) should have been close to that of the car body (anchor), but in fact it doesn't.
When using the original triplet loss, the depth prediction of the background negatives is so good ($\mathcal{D}^-$ is so large enough) that they can shelter / cover up the error of the positives.
In consequence, Eq.~\ref{eq:expect} is satisfied and no learning happens - all poor positive pixels (car window) are not going to be optimized.

\subsubsection{Solution 2: Optimize Pos and Neg in Isolation}

To this end, we no longer compare the value of $(\mathcal{D}^--\mathcal{D}^+)$ with $m$.
Instead, we easily split $\mathcal{D}^+$ and $\mathcal{D}^-$ from within the original triplet loss (Eq.~\ref{eq:base_triplet}).
We compare $\mathcal{D}^-$ with a new margin $m^{\prime}$, and optimize $\mathcal{D}^+$ directly.
Concretely, our redesigned the triplet loss becomes:
\begin{equation}\label{eq:contrastive}
    \mathcal{L}_{tri}^{\prime} = \frac{1}{|\Gamma|} \sum_{\mathcal{P}_i \in \Gamma} \left( \mathcal{D}^{+}(i) + \left[ m^{\prime} - \mathcal{D}^{-\prime}(i) \right]_{+} \right),
\end{equation}
where either error of the positives or negatives appears individually without any mutual effects.
\begin{table*}[bp]
    \scriptsize
    \begin{tabular}{ l | c | c | c || c | c | c | c || c | c | c }
        \Xhline{0.8pt} Method & PP & W $\times$ H & Data & \cellcolor{pink}Abs Rel    & \cellcolor{pink}Sq Rel    & \cellcolor{pink}RMSE   &  \cellcolor{pink}\begin{tabular}[c]{@{}c@{}}RMSE \\ log \end{tabular} & \cellcolor{SkyBlue2}$\delta \textless 1.25$   & \cellcolor{SkyBlue2}$\delta \textless 1.25^2$   & \cellcolor{SkyBlue2}$\delta \textless 1.25^3$ \\

        \hline Ranjan \textit{et al.}~\cite{ranjan2019competitive} & \textcolor{X_color}{\XSolidBrush} & $832 \times 256$ & S & 0.148 & 1.149 & 5.464 & 0.226 & 0.815 & 0.935 & 0.973 \\
        EPC++~\cite{luo2019every} & \textcolor{X_color}{\XSolidBrush} & $832 \times 256$ & S & 0.141 & 1.029 & 5.350 & 0.216 & 0.816 & 0.941 & 0.976 \\
        Structure2depth~\cite{casser2019depth} & \textcolor{X_color}{\XSolidBrush} & $416 \times 128 $ & M & 0.141 & 1.026 & 5.291 & 0.215 & 0.816 & 0.945 & 0.979 \\
        Videos in the wild~\cite{gordon2019depth} & \textcolor{X_color}{\XSolidBrush} & $416 \times 128 $ & M & 0.128 & 0.959 & 5.230 & 0.212 & 0.845 & 0.947 & 0.976 \\
        Guizilini \textit{et al.}~\cite{guizilini2020semantically} & \textcolor{X_color}{\XSolidBrush} & $640 \times 192 $ & M & 0.102 & \underline{0.698} & \underline{4.381} & 0.178 & 0.896 & 0.964 & \textbf{0.984} \\
        Johnston \textit{et al.}~\cite{johnston2020self} & \textcolor{X_color}{\XSolidBrush} & $640 \times 192 $ & M & 0.106 & 0.861 & 4.699 & 0.185 & 0.889 & 0.962 & 0.982 \\
        Packnet-SFM~\cite{guizilini20203d} & \textcolor{X_color}{\XSolidBrush} & $640 \times 192 $ & M & 0.111 & 0.785 & 4.601 & 0.189 & 0.878 & 0.960 & 0.982 \\
        Li \textit{et al.}~\cite{li2020unsupervised} & \textcolor{X_color}{\XSolidBrush} & $416 \times 128 $ & M & 0.130 & 0.950 & 5.138 & 0.209 & 0.843 & 0.948 & 0.978 \\
        Patil \textit{et al.}~\cite{patil2020don} & \textcolor{X_color}{\XSolidBrush} & $640 \times 192 $ & M & 0.111 & 0.821 & 4.650 & 0.187 & 0.883 & 0.961 & 0.982 \\
        Wang \textit{et al.}~\cite{wang2020self} & \textcolor{X_color}{\XSolidBrush} & $640 \times 192 $ & M & 0.106 & 0.799 & 4.662 & 0.187 & 0.889 & 0.961 & 0.982 \\

        Monodepth2 MS~\cite{godard2019digging}  & \textcolor{X_color}{\XSolidBrush} & $640 \times 192$ & MS & 0.106  & 0.818  & 4.750  & 0.196 & 0.874 & 0.957 &  0.979  \\

        Zhou \textit{et al.}~\cite{zhou2017unsupervised} & \textcolor{X_color}{\XSolidBrush} & $640 \times 192$ & M & 0.183 & 1.595 & 6.709 & 0.270 & 0.734 & 0.902 & 0.959 \\

        WaveletMonodepth~\cite{ramamonjisoa2021single}  & \textcolor{X_color}{\XSolidBrush} & $640 \times 192$ & S & 0.109 &  0.845 & 4.800  & 0.196 & 0.870 & 0.956 & 0.980  \\

        HR-Depth~\cite{lyu2020hr} & \textcolor{X_color}{\XSolidBrush} & $640 \times 192$ & MS & 0.107 & 0.785 & 4.612 & 0.185 & 0.887 & 0.962 & 0.982  \\

        FSRE-Depth~\cite{jung2021fine}  & \textcolor{X_color}{\XSolidBrush} & $640 \times 192$ & M & 0.105 & 0.722 & 4.547 & 0.182 & 0.886 & 0.964 & 0.984  \\

        Depth-Hints~\cite{watson2019self}  & \textcolor{X_color}{\XSolidBrush} & $640 \times 192$ & S & 0.109 & 0.845 & 4.800 & 0.196 & 0.870 & 0.956 & 0.980  \\

        CADepth~\cite{yan2021channel}  & \textcolor{X_color}{\XSolidBrush} & $640 \times 192$ & S & 0.106 & 0.849 & 4.885 & 0.204 & 0.869 & 0.951 & 0.976  \\

        SuperDepth~\cite{pillai2019superdepth}  & \textcolor{X_color}{\XSolidBrush} & $640 \times 192$ & S &  0.112 & 0.875 & 4.958 & 0.207 & 0.852 & 0.947 & 0.977  \\

        ManyDepth~\cite{watson2021temporal}  & \textcolor{X_color}{\XSolidBrush} & $640 \times 192$ & M & \underline{0.098} & 0.770 & 4.459 & \underline{0.176} & \underline{0.900} & \underline{0.965} & \underline{0.983}  \\

        Refine\&Distill~\cite{pilzer2019refine}  & \textcolor{X_color}{\XSolidBrush} & -- & S &  \underline{0.098} & 0.831 & 4.656 & 0.202 & 0.882 & 0.948 & 0.973  \\

        \textbf{TriDepth (Ours)} & \textcolor{X_color}{\XSolidBrush} & $640 \times 192$ & M & \textbf{0.093} & \textbf{0.665} & \textbf{4.272} & \textbf{0.172} & \textbf{0.907} & \textbf{0.967} & \textbf{0.984} \\
        \Xhline{0.8pt}
    \end{tabular}
    \vspace{0.12cm}
    \centering
    \caption{\textbf{Comparison to previous SoTA on KITTI Eigen split~\cite{eigen2014depth}.} Metrics are \colorbox{pink}{error metrics \textdownarrow} and \colorbox{SkyBlue2}{accuracy metrics \textuparrow}. The \textit{Data} column specifies the training data type: S - stereo images, M - monocular video and MS - stereo video. All models are trained with $192\times 640$ images and Resnet18~\cite{he2016deep} as backbone. All results are not \textbf{P}ost-\textbf{P}rocessed~\cite{godard2017unsupervised}. Best results are in \textbf{bold}; second best are \underline{underlined}. We outperform all previous methods by a large margin on exactly all metrics.}
    \label{tab:kitti_res}
\end{table*}
\textit{E.g.}, in Fig.~\ref{fig:need-of-contrastive}, the poor positives (wrong car window depth predictions) will be penalized directly, since $\mathcal{D}^+$ is no longer restricted to the comparison with $\mathcal{D}^-$.
This is a bit like contrastive loss~\cite{hadsell2006dimensionality}, but we point out that we are still a triplet loss, since we still follow the idea of searching and optimizing both positives and negatives of the given anchor simultaneously.
Another reason why we do not optimize the value of $(\mathcal{D}^- - \mathcal{D}^+)$ is that, we have no prior knowledge of how much the depth difference between the two-side objects should be.
In fact, it varies with different objects.
But one thing is certain: the smaller the $\mathcal{D}^+$, the better, since we always want the depths of different parts of the same object to be the same.

\section{Experiments}\label{sec:experiments}

\subsection{Implementation Details}\label{subsec:implementation-details}

We implement our methods in PyTorch~\cite{paszke2017automatic}.
We set the total training epoch as 20, the learning rate as $10^{-4}$.
Our triplet loss is applied throughout every layer of the depth decoder.
We set the triplets' patch size to be $5\times 5$, and $k=4$~\cite{jung2021fine} to filter out anchors located in potentially inaccurate semantic predictions.
Because we use the minimum $\mathcal{D}^-$ of all negative samples, we increase the original margin $m=0.3$ to $m^{\prime}=0.65$.
The rest of our model's setting is the same as~\cite{watson2021temporal}.
Note that our triplet loss is not restricted to a fixed baseline, when plugging it into a new model, just leave all the original settings exactly unchanged.

\begin{table*}[t]
    \scriptsize
    \begin{tabular}{ l | c | c | c | c || c | c | c | c | c || c | c | c }
        \Xhline{0.8pt} Method & Pub. & PP & W $\times$ H & Data & \cellcolor{pink}\begin{tabular}[c]{@{}c@{}}Extra \\ time \end{tabular} & \cellcolor{pink}Abs Rel    & \cellcolor{pink}Sq Rel  & \cellcolor{pink}RMSE   &  \cellcolor{pink}\begin{tabular}[c]{@{}c@{}}RMSE \\ log \end{tabular} & \cellcolor{SkyBlue2}$\delta_1$   & \cellcolor{SkyBlue2}$\delta_2$   & \cellcolor{SkyBlue2}$\delta_3$ \\
        \hline
        \specialrule{0em}{1pt}{1pt}

        \hline Monodepth2 M~\cite{godard2019digging} & ICCV 2019 & \textcolor{X_color}{\XSolidBrush} & $640 \times 192$ & M & -- & 0.115   & 0.903  & 4.863  & 0.193  & 0.877  & 0.959  & 0.981  \\
        \cellcolor{aug_our_method_color}\textbf{+ Ours} & -- & \textcolor{X_color}{\XSolidBrush} & $640 \times 192$ & M & \textbf{+ 0ms} & \textbf{0.108}  & \textbf{0.744} & \textbf{4.537} & \textbf{0.184} & \textbf{0.883} & \textbf{0.963} & \textbf{0.983}  \\
        \hline \specialrule{0em}{1pt}{1pt}

        \hline Zhou \textit{et al.}~\cite{zhou2017unsupervised} & CVPR 2017 & \textcolor{X_color}{\XSolidBrush} & $640 \times 192$ & M & -- & 0.183 & 1.595 & 6.709 & 0.270 & 0.734 & 0.902 & 0.959 \\
        \cellcolor{aug_our_method_color}\textbf{+ Ours} & -- & \textcolor{X_color}{\XSolidBrush} & $640 \times 192$ & M & \textbf{+ 0ms} & \textbf{0.148}  &  \textbf{1.098}  &  \textbf{5.150}  &  \textbf{0.212}  &  \textbf{0.819}  &  \textbf{0.949}  &  \textbf{0.980}  \\
        \hline \specialrule{0em}{1pt}{1pt}

        \hline Monodepth2 S~\cite{godard2019digging} & ICCV 2019 & \textcolor{X_color}{\XSolidBrush} & $640 \times 192$ & S & -- & 0.109   & 0.873  & 4.960  & 0.209  & 0.864  & 0.948  & 0.975  \\
        \cellcolor{aug_our_method_color}\textbf{+ Ours} & -- & \textcolor{X_color}{\XSolidBrush} & $640 \times 192$ & S & \textbf{+ 0ms} & \textbf{0.107} & \textbf{0.826} & \textbf{4.822} & \textbf{0.201} & \textbf{0.866}  & \textbf{0.953} & \textbf{0.978}  \\
        \hline \specialrule{0em}{1pt}{1pt}

        \hline FSRE-Depth only SGT~\cite{jung2021fine} & ICCV 2021 & \textcolor{X_color}{\XSolidBrush} & $640 \times 192$ & M & -- & 0.113 & 0.836 & 4.711 & 0.187 & 0.878 & 0.960 & 0.982 \\
        \cellcolor{aug_our_method_color}\textbf{+ Ours} & -- & \textcolor{X_color}{\XSolidBrush} & $640 \times 192$ & M & \textbf{+ 0ms} & \textbf{0.108} &  \textbf{0.746}  &  \textbf{4.507} &  \textbf{0.182} & \textbf{0.884} & \textbf{0.964} &  \textbf{0.983} \\
        \hline \specialrule{0em}{1pt}{1pt}

        \hline Monodepth2 MS~\cite{godard2019digging} & ICCV 2019 & \textcolor{X_color}{\XSolidBrush} & $640 \times 192$ & MS & -- & 0.106  & 0.818  & 4.750  & 0.196 & 0.874 & 0.957 &  0.979  \\
        \cellcolor{aug_our_method_color}\textbf{+ Ours} & -- & \textcolor{X_color}{\XSolidBrush} & $640 \times 192$ & MS & \textbf{+ 0ms} & \textbf{0.105} & \textbf{0.753} & \textbf{4.563} & \textbf{0.182} & \textbf{0.887} & \textbf{0.963} & \textbf{0.983}  \\
        \hline \specialrule{0em}{1pt}{1pt}

        \hline Depth-Hints~\cite{watson2019self} & ICCV 2019 & \textcolor{X_color}{\XSolidBrush} & $640 \times 192$ & S & -- & 0.109 &  0.845 & 4.800  & 0.196 & 0.870 & 0.956 &
        \textbf{0.980} \\
        \cellcolor{aug_our_method_color}\textbf{+ Ours} & -- & \textcolor{X_color}{\XSolidBrush} & $640 \times 192$ & S & \textbf{+ 0ms} & \textbf{0.106} & \textbf{0.843} & \textbf{4.774} & \textbf{0.194} & \textbf{0.875} & \textbf{0.957} & \textbf{0.980} \\
        \hline \specialrule{0em}{1pt}{1pt}

        \hline HR-Depth~\cite{lyu2020hr} & AAAI 2020 & \textcolor{X_color}{\XSolidBrush} & $640 \times 192$ & M & -- & 0.109 & 0.792 & 4.632 & 0.185 & 0.884 & 0.962 & 0.983 \\
        \cellcolor{aug_our_method_color}\textbf{+ Ours} & -- & \textcolor{X_color}{\XSolidBrush} & $640 \times 192$ & M & \textbf{+ 0ms} & \textbf{0.107} & \textbf{0.760} & \textbf{4.522} & \textbf{0.182} & \textbf{0.886} & \textbf{0.964} & \textbf{0.984} \\
        \hline \specialrule{0em}{1pt}{1pt}

        \hline HR-Depth MS~\cite{lyu2020hr} & AAAI 2020 & \textcolor{X_color}{\XSolidBrush} & $640 \times 192$ & MS & -- & 0.107 & 0.785 & 4.612 & 0.185 & 0.887 & 0.962 & 0.982 \\
        \cellcolor{aug_our_method_color}\textbf{+ Ours} & -- & \textcolor{X_color}{\XSolidBrush} & $640 \times 192$ & MS & \textbf{+ 0ms} & \textbf{0.105} & \textbf{0.751} & \textbf{4.512} & \textbf{0.181} & \textbf{0.890} & \textbf{0.963} & \textbf{0.983} \\
        \hline \specialrule{0em}{1pt}{1pt}

        \hline ManyDepth~\cite{watson2021temporal} & CVPR 2021 & \textcolor{X_color}{\XSolidBrush} & $640 \times 192$ & M & -- & 0.098 & 0.770 & 4.459 & 0.176 & 0.900 & 0.965 & 0.983  \\
        \cellcolor{aug_our_method_color}\textbf{+ Ours} & -- & \textcolor{X_color}{\XSolidBrush} & $640 \times 192$ & M & \textbf{+ 0ms} & \textbf{0.093} & \textbf{0.665} & \textbf{4.272} & \textbf{0.171} & \textbf{0.907} & \textbf{0.967} & \textbf{0.984} \\
        \hline \specialrule{0em}{1pt}{1pt}

        \hline CADepth~\cite{yan2021channel} & 3DV 2021 & \textcolor{X_color}{\XSolidBrush} & $640 \times 192$ & M & -- & 0.110 & 0.812 & 4.686 & 0.187 & 0.882 & 0.962 & 0.983 \\
        \cellcolor{aug_our_method_color}\textbf{+ Ours} & -- & \textcolor{X_color}{\XSolidBrush} & $640 \times 192$ & M & \textbf{+ 0ms} & \textbf{0.105} & \textbf{0.745} & \textbf{4.530} & \textbf{0.181} & \textbf{0.888} & \textbf{0.965} & \textbf{0.984} \\

        \Xhline{0.8pt}
    \end{tabular}
    \vspace{0.2cm}
    \centering
    \caption{\textbf{Comparisons of existing models with and without our method on KITTI Eigen split~\cite{eigen2014depth}.} All models are trained with $192\times 640$ images and Resnet18~\cite{he2016deep} as backbone. All results are not \textbf{P}ost-\textbf{P}rocessed~\cite{godard2017unsupervised}. Models \colorbox{aug_our_method_color}{augmented with our powerful triplet loss} are highlighted, achieving better results than their original counterparts on exactly all metrics, while no extra inference computation is needed.}
    \label{tab:generalization}
\end{table*}

\definecolor{emphasize_line_color}{RGB}{241,199,245}
\begin{table*}[htb]
    \scriptsize
    \begin{tabular}{ c | c | c || c | c | c | c || c | c | c }
        \Xhline{0.8pt}
        \begin{tabular}[c]{@{}c@{}}Naive \\Triplet Loss \end{tabular} & \begin{tabular}[c]{@{}c@{}}Hard \\Negative \end{tabular} & \begin{tabular}[c]{@{}c@{}}Isolated \\Triplet \end{tabular} & \cellcolor{pink}Abs Rel & \cellcolor{pink}Sq Rel    & \cellcolor{pink}RMSE   &  \cellcolor{pink}\begin{tabular}[c]{@{}c@{}}RMSE \\ log \end{tabular} & \cellcolor{SkyBlue2}$\delta_1$   & \cellcolor{SkyBlue2}$\delta_2$   & \cellcolor{SkyBlue2}$\delta_3$ \\
        \hline

        \arrayrulecolor{shallow_grey}\hline
        & & & 0.115 & 0.903 & 4.863 & 0.193 & 0.877 & 0.959 & 0.981 \\

        \arrayrulecolor{shallow_grey}\hline
        \textcolor{checkmark_color}{\checkmark} & & & 0.113 & 0.836 & 4.711 & 0.187 & 0.878 & 0.960 & 0.982 \\

        \arrayrulecolor{shallow_grey}\hline
        \textcolor{checkmark_color}{\checkmark} & \textcolor{checkmark_color}{\checkmark} & & 0.110 &  0.782  &  4.622  &  0.185  &  0.881  &  0.963  &  \textbf{0.983} \\

        \arrayrulecolor{shallow_grey}\hline
        \textcolor{checkmark_color}{\checkmark} & & \textcolor{checkmark_color}{\checkmark} & 0.111  &  0.823  &  4.645  &  0.186  &  0.881  &  0.962  &  0.982 \\

        \arrayrulecolor{shallow_grey}\hline
        \textcolor{checkmark_color}{\checkmark} & \textcolor{checkmark_color}{\checkmark} & \textcolor{checkmark_color}{\checkmark} & \textbf{0.108}  &  \textbf{0.746}  &  \textbf{4.507}  &  \textbf{0.182}  &  \textbf{0.884}  &\textbf{0.964} &  \textbf{0.983} \\

        \Xhline{0.8pt}
    \end{tabular}
    \vspace{0.2cm}
    \centering
    \caption{\textbf{Ablations of our different strategies on KITTI Eigen split~\cite{eigen2014depth}.} Here, we use \textit{Monodepth2}~\cite{godard2019digging} as the baseline. All models are trained with $192\times 640$ monocular videos and Resnet18~\cite{he2016deep} as backbone. All results are not post-processed~\cite{godard2017unsupervised}.}
    \label{tab:abalation}
\end{table*}

\subsection{Comparison to State-of-the-Arts}\label{subsec:quantitative-results}

\begin{figure}[htb]
    \centering\includegraphics[width=\columnwidth]{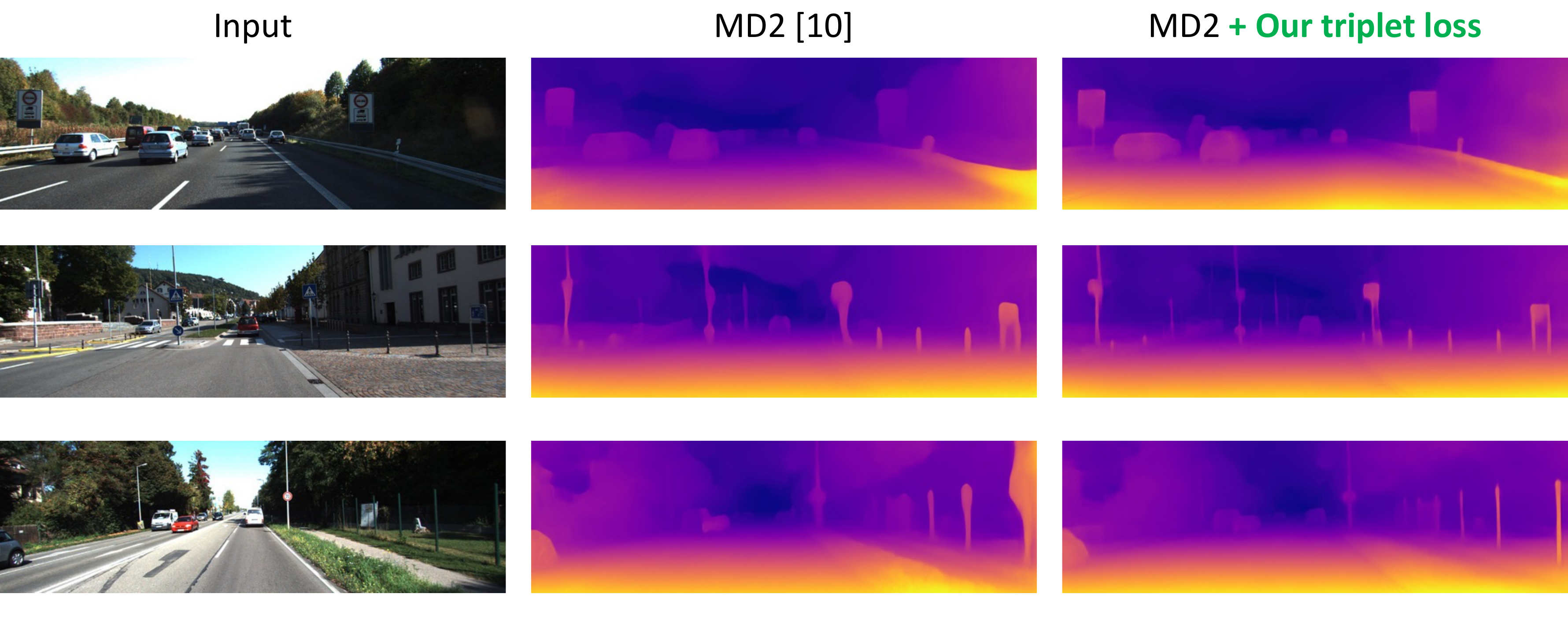}
    \caption{\textbf{Visual comparisons of an existing model with and w/o our triplet loss.} The edge-fattening problem is significantly alleviated. The depths of thin structures are better aligned to their RGB counterparts, \textit{e.g.} road signs and poles.}
    \label{fig:depth_preds}
\end{figure}

KITTI dataset~\cite{geiger2012we} is selected as our testing benchmark where calibrated stereo images with depth information from the Lidar point clouds are provided.
In detail, we use the KITTI Eigen split~\cite{eigen2014depth} to evaluate the depth predictions, which is commonly used for comparisons of self-supervised MDE models.
The depth ground truth is from the provided sparse Lidar point clouds, which is used to measure all seven standard metrics.
As for fair comparison, we adopt the standard depth ceiling of 80 meters~\cite{garg2016unsupervised}.
Please refer to~\cite{eigen2014depth} for more evaluation details.
We report the results in Tab.~\ref{tab:kitti_res}, showing that we produce an unprecedented performance.
Specifically, even though we make no use of any stereo pairs during training, our model still outperforms all SoTA by a large margin on every metric.
To enable real-time applications, we abandon the post-processing technique~\cite{godard2019digging} which doubles the inference computation, even so, we still show comparable results to previous post-processed SoTA results~\cite{watson2019self,lyu2020hr,jung2021fine,yan2021channel}.
Visualization of solving the edge-fattening issue is shown in Fig.~\ref{fig:depth_preds}.

\subsection{Generalizability}\label{subsec:generalizability}
As mentioned before, our triplet loss is not only powerful, but also versatile and lightweight.
After showing our unparalleled state-of-the-art performance, we demonstrate the high generalizability of the proposed triplet loss.

We integrate our triplet loss into a large number of existing open-source methods (some of them are SoTA before ours), and show that our triplet loss can help them to achieve better scores, as reported in Tab.~\ref{tab:generalization}.
Remarkably, in all integrations, we bring a substantial performance boost to the original model, in exactly all seven metrics.
Furthermore, we do not lead in any extra inference computation at all, which shows our lightness.
For the benefit of the whole MDE community, we publish our full implementation, since we believe this powerful triplet loss could enhance more subsequent depth estimators.

\subsection{Ablation Study}\label{subsec:ablation-study}

In Tab.~\ref{tab:abalation}, we validate the effectiveness of our two design decisions in turn.
The original triplet loss, migrated from deep metric learning by Jung \textit{et al.}~\cite{jung2021fine}, yields only a slight improvement to the baseline model~\cite{godard2019digging}.
Through either of our two contributions, the performance is substantially improved, and not surprisingly, our full model performs the best.
It is also worth noting that none of our contributions introduces any computation overhead when inferring.

\subsubsection{Consideration on Margin $m^{\prime}$}
We further compare different values of the margin $m^{\prime}$ in our redesigned triplet loss, as shown in Tab.~\ref{tab:abalation_on_margin}.
We select $m^{\prime}=0.65$ because it shows the best result.
Either increasing or decreasing $m^{\prime}$ leads to performance degradation.
We point out that selecting the margin of triplet loss in depth estimation is a difficult task, since we have no prior knowledge of whether the semantic boundaries guarantee a large depth difference or not.
When $m^{\prime}$ is too large, the triplet loss would be too strict, \textit{i.e.} leading to some false positives. \textit{E.g.}, let us consider an $m^{\prime}$ encouraging the depth difference between the intersecting objects to be $100$m, then, the boundaries with a depth difference of $5$m will be wrongly penalized.
When $m^{\prime}$ is too small, the triplet loss will be too tolerant, \textit{i.e.} leading to some false negatives. \textit{E.g.}, let us consider an $m^{\prime}$ encouraging the depth difference between the intersecting objects just to be $5$m, then, boundaries with a depth difference of $100$m but the prediction only has a depth difference of $80$m, are going to be neglected by the optimization strategy.

\begin{table}[htb]
    \scriptsize
    \begin{tabular}{ c || c | c | c | c || c | c | c }
        \Xhline{0.8pt}
        $m^{\prime}$ & \cellcolor{pink}Abs Rel & \cellcolor{pink}Sq Rel    & \cellcolor{pink}RMSE   &  \cellcolor{pink}R log & \cellcolor{SkyBlue2}$\delta_1$   & \cellcolor{SkyBlue2}$\delta_2$   & \cellcolor{SkyBlue2}$\delta_3$ \\
        \hline
        0.50 & 0.110 & 0.818 & 4.646 &  0.186 & 0.882 & 0.962 & 0.982 \\

        0.60 & 0.109  &  0.806  &  4.667 & 0.184 & 0.883 &  0.961 & 0.983 \\

        0.65 & \textbf{0.108} & \textbf{0.746}  &  \textbf{4.507}  &  \textbf{0.182}  &  \textbf{0.884}  &\textbf{0.964} &  \textbf{0.983} \\

        0.70 & 0.109 & 0.763 & 4.613 & 0.184 & 0.882 & 0.963 &  0.983 \\

        0.80 & 0.110 & 0.787 & 4.601 & 0.185 & 0.880 & 0.962 &  0.983 \\
        \Xhline{0.8pt}
    \end{tabular}
    \vspace{0.12cm}
    \centering
    \caption{\textbf{Ablation on margin $m^{\prime}$.} All models are trained with $192\times 640$ monocular videos and Resnet18~\cite{he2016deep} as backbone. All results are not post-processed~\cite{godard2017unsupervised}.}
    \label{tab:abalation_on_margin}
\end{table}

\subsection{Can We Trust in Current MDE?}\label{subsec:discussion}

Apart from solving the ubiquitous edge-fattening problem, we also reveal new problems for our whole MDE community as the future works.
We hope our community could focus not only on the performance numbers, but also on the barriers that prevent our practical application.

\begin{figure}
    \centering
    \centerline{\includegraphics[width=\columnwidth]{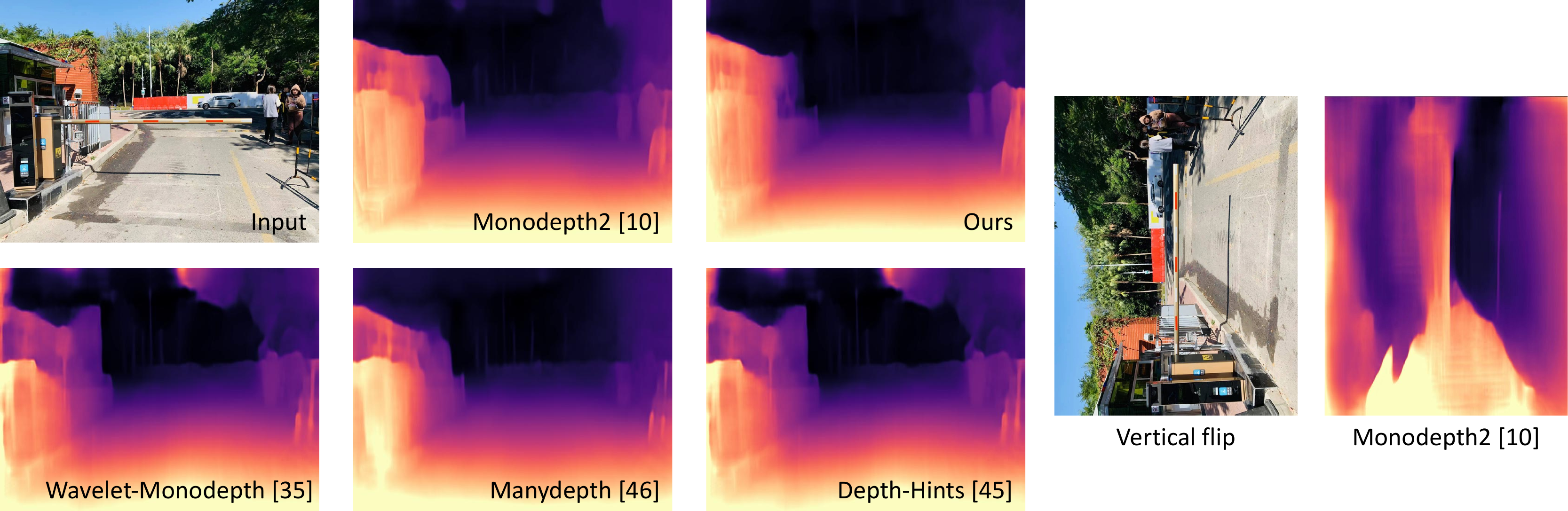}}
    \caption{
        \textbf{A fatal failure case for autonomous driving.} The depth predictions of the crossbar in the middle of the input image are totally wrong, which may lead to a fatal error in autonomous driving. When vertically flipping the input image, the depth of the crossbar is correctly estimated, while other objects, \textit{e.g.}, people, are wrong again.
    }
    \label{fig:failure_case}
\end{figure}

The scenario shown in Fig.~\ref{fig:failure_case} is not from any open datasets, but is captured by one of our authors.
As can be seen, quantities of self-supervised MDE models fail completely in predicting the depth of the crossbar.
We speculate that there are two probable reasons:
\vspace{-0.1cm}
\begin{itemize}
    \item Existing methods rely too heavily on the images' shape features in the dataset, and by chance, there are no horizontal bars in KITTI~\cite{geiger2012we} dataset.
    \item When training with stereo pairs, for the horizontal bar, the photometric loss for all disparities is small, thus no clear global optimum exists.
    The network therefore tends to mix the crossbar and the background road together.
\end{itemize}

\vspace{-0.1cm}
We expect future MDE models to better generalize to real-life scenes via better feature representation techniques, \textit{e.g.} deformable convolution~\cite{Dai_2017_ICCV} or data augmentation in feature space~\cite{devries2017dataset}.

\section{Conclusion}\label{sec:conclusion}

In this paper, we solve the notorious and ubiquitous edge-fattening issue in self-supervised MDE by introducing a well-designed triplet loss.
We tackle two drawbacks of the original patch-based triplet loss in MDE.
First, we propose to apply \textit{min.} operator on computing anchor-negative distance, to prevent the error of the edge-fattening negatives from being masked.
Second, we split the anchor-positive distance and anchor-negative distance from within the original triplet loss.
This strategy provides more direct optimizations to the positives, without the mutual effect with the negatives.
Our triplet loss is highly powerful, versatile and lightweight.
Experiments show that it not only brings our \textit{TriDepth} model an unprecedented SoTA performance, but also provides substantial performance boosts to a large number of existing models, without introducing any extra inference computation at all.

\vspace{0.2cm}
\noindent\textbf{Acknowledgement.}
This work was supported by National Natural Science Foundation of China (No. 62172021).

{\small
\bibliographystyle{ieee_fullname}
\bibliography{main}
}

\end{document}